\documentclass{article}
\usepackage{spconf,amsmath,graphicx}
\usepackage{amssymb}
\usepackage{array}
\usepackage{float}
\usepackage{dblfloatfix}
\usepackage{xcolor}
\usepackage{colortbl}
\usepackage{enumitem}

\title{Adaptive Multi-Scale Correlation Meta-Network for Few-Shot Remote Sensing Image Classification}

\name{
\parbox{\linewidth}{
\centering
Anurag Kaushish$^\dagger$ \quad
Ayan Sar$^\dagger$ \quad
Sampurna Roy$^\dagger$ \quad
Sudeshna Chakraborty$^\ddagger$ \\
Prashant Trivedi$^\ast$ \quad
Tanupriya Choudhury$^\dagger$ \quad
Kanav Gupta$^\dagger$
}
}

\address{
$^{\dagger}$ School of Computer Science, UPES, Dehradun, 248007, Uttarakhand, India.
\\
$^{\ddagger}$ School of Computer Science and Engineering, Galgotias University, India.
\\
$^{\ast}$ Computer Science and Information Systems, BITS Pilani, Pilani Rajasthan, India.
}

\begin{document}
%
\maketitle
\begin{abstract}
Few-shot learning in remote sensing remains challenging due to three factors: the scarcity of labeled data, substantial domain shifts, and the multi-scale nature of geospatial objects. To address these issues, we introduce Adaptive Multi-Scale Correlation Meta-Network (AMC-MetaNet), a lightweight yet powerful framework with three key innovations: (i) correlation-guided feature pyramids for capturing scale-invariant patterns, (ii) an adaptive channel correlation module (ACCM) for learning dynamic cross-scale relationships, and (iii) correlation-guided meta-learning that leverages correlation patterns instead of conventional prototype averaging. Unlike prior approaches that rely on heavy pre-trained models or transformers, AMC-MetaNet is trained from scratch with only $\sim600K$ parameters, offering $20\times$ fewer parameters than ResNet-18 while maintaining high efficiency ($<50$ms per image inference). AMC-MetaNet achieves up to 86.65\% accuracy in 5-way 5-shot classification on various remote sensing datasets, including EuroSAT, NWPU-RESISC45, UC Merced Land Use, and AID. Our results establish AMC-MetaNet as a computationally efficient, scale-aware framework for real-world few-shot remote sensing.
\end{abstract}
\begin{keywords}
Few-Shot Classification, Remote Sensing, Meta-Learning, Scale-Invariant Representation
\end{keywords}

\section{Introduction} \label{sec:intro}
Remote sensing has emerged as a critical enabler for environmental monitoring, urban planning, disaster management, and agricultural assessment. The increasing availability of high-resolution satellite imagery has opened new avenues for automated interpretation; however, effective classification of remote sensing data remains challenging  \cite{Zhang_2025}. One of the most pressing issues is the scarcity of labelled data, especially in domains where expert annotations are costly and time-consuming. This makes few-shot learning (FSL) \cite{Dang_2023} an essential research direction, enabling models to generalize from only a handful of annotated examples.

\begin{figure}[t!]
    \centering
    \includegraphics[width=\linewidth]{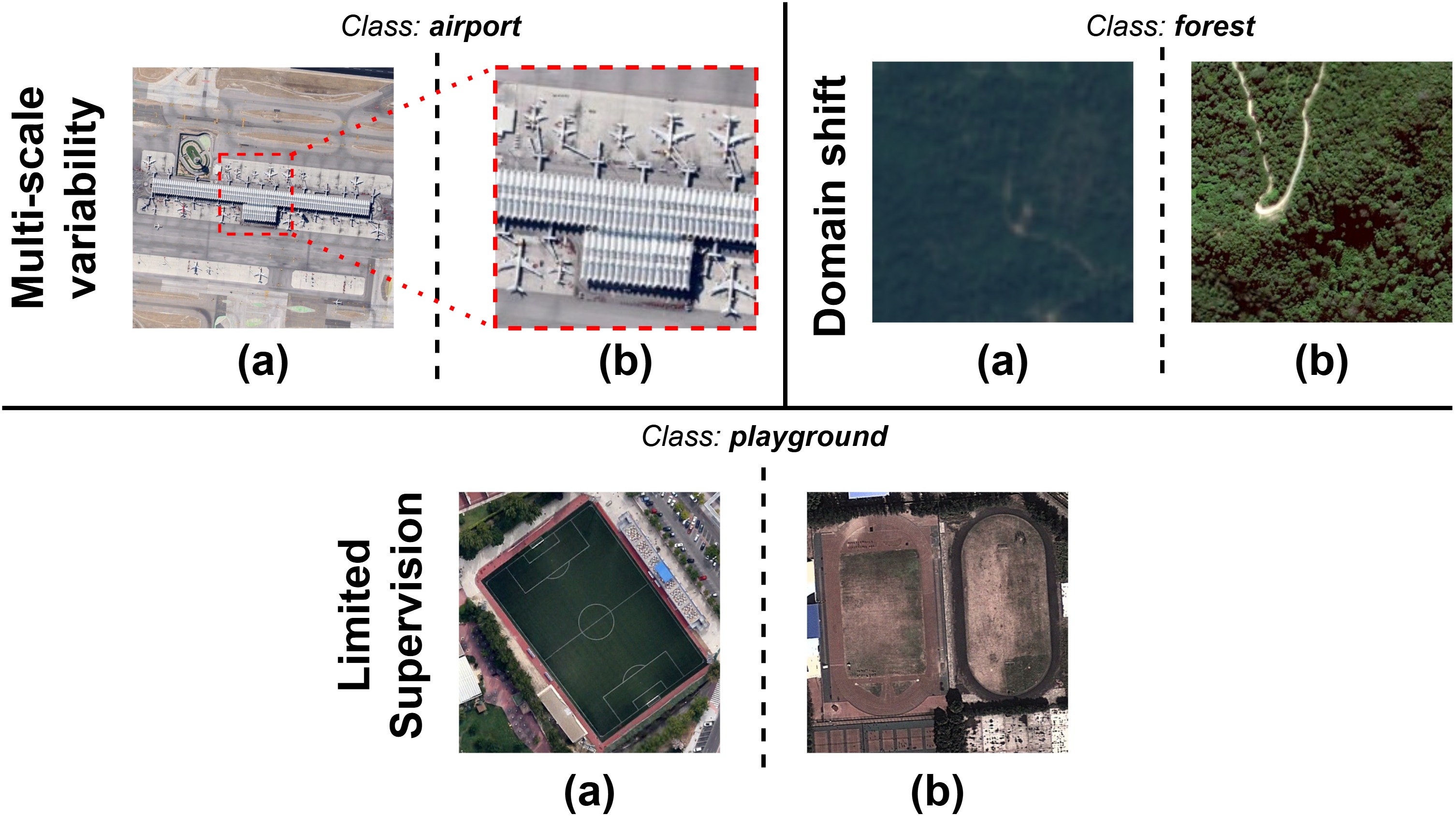}
    \caption{Examples illustrating key challenges in remote sensing (RS) image classification: (Row 1) (Left) NWPU dataset showing an airplane from a wider view (a) and a zoomed view (b), highlighting the multi-scale challenge; (Row 1) (Right) EuroSAT (a) and AID (b) datasets showing the "forest" class, demonstrating domain shift where models trained on one dataset may fail to classify visually different instances from another; (Row 2) AID dataset with a soccer field labeled as "playground" (a) and another playground with a running track (b), showing the difficulty of generalizing from limited samples to visually distinct instances of the same class.}
    \label{fig:rsinto}
\end{figure}

\begin{figure*}[t!]
    \centering
    \includegraphics[width=\textwidth]{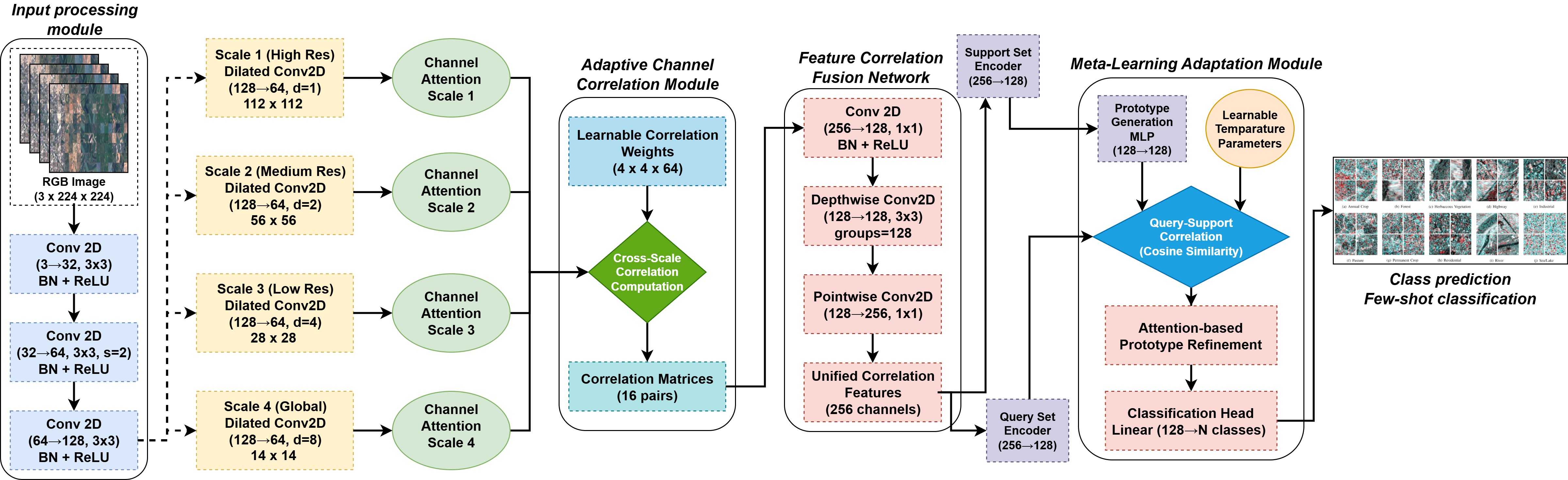}
    \caption{Architecture of the proposed AMC-MetaNet.}
    \label{fig:arch}
\end{figure*}

Traditional deep learning models, despite their success in natural image classification, struggle in the remote sensing domain due to three fundamental challenges: (i) multi-scale variability where objects appear at vastly different resolutions, (ii) domain shift across different sensors and geographical regions \cite{Aswal_2024}, and (iii) limited supervision in realistic scenarios (see Fig. \ref{fig:rsinto}). Existing FSL (Few-Shot Learning) solutions often rely on pre-trained backbones or transformer-based architectures, which are computationally heavy and may fail to capture the unique scale-dependent structures of remote sensing imagery \cite{Sun_2021}.

To address these challenges, we propose a novel method, AMC-MetaNet (Adaptive Multi-Scale Correlation Meta-Network), specifically designed for few-shot remote sensing classification. The key idea is to replace conventional feature concatenation with correlation-guided multi-scale processing, enabling robust adaptation without pre-training dependencies. By integrating cross-scale correlation learning and meta-adaptation, AMC-MetaNet provides an efficient, domain-tailored approach for FSL in remote sensing (see Fig. \ref{fig:arch}). The key novel contributions of this work are:
\begin{itemize}[noitemsep, topsep=2pt]
    \item \textbf{Multi-Scale Correlation-Guided Features:} We introduce a multi-scale feature extraction strategy that computes learnable correlations across pyramid levels, capturing scale-invariant patterns crucial for remote sensing tasks.
    \item \textbf{Adaptive Channel Correlation Module (ACCM):} We designed a novel module with learnable correlation weights and channel attention mechanisms, dynamically modeling cross-scale dependencies without external pre-trained models.
    \item \textbf{Correlation-Guided Meta-Learning:}  We proposed a meta-adaptation strategy based on correlation patterns rather than simple prototype averaging, enhancing few-shot adaptability under domain shift conditions.
\end{itemize}
Together, these innovations establish AMC-MetaNet as a modular, scale-aware, and computationally efficient framework that advances few-shot learning in remote sensing.
\section{Methodology} \label{methodology}
\subsection{Problem Formulation}

We define the few-shot remote sensing classification task under the standard episodic learning framework. Each episode consists of a support set $\mathcal{S} = \{(x_i^s, y_i^s)\}_{i=1}^{N \times K}$ and a query set $\mathcal{Q} = \{(x_j^q, y_j^q)\}_{j=1}^{N \times Q}$, where $N$ is the number of classes, $K$ is the number of support examples per class (shot), and $Q$ is the number of queries. We use superscript $s$ to denote support and $q$ to represent the query. The objective is to learn a classifier $f_{\theta}$ that generalizes to unseen classes with minimal supervision, i.e., $f_{\theta}: \mathcal{S}, \mathcal{Q} \mapsto \hat{y}_j^q$, where $\theta$ denotes the learnable parameters. We now present the AMC-MetaNet architecture, a novel framework designed to achieve this goal.

\subsection{Overall Architecture of AMC-MetaNet}
Given an input remote sensing image $x \in \mathbb{R}^{3 \times H \times W}$, the AMC-MetaNet processing pipeline proceeds as follows:  
\begin{itemize}[noitemsep, topsep=2pt]
    \item \textbf{Initial feature extraction:}  
    The input image is passed through a lightweight convolutional network to obtain a compact yet discriminative feature map  $F \in \mathbb{R}^{C \times h \times w}$,
    where $C$ denotes the number of channels and $(h, w)$ are the reduced spatial dimensions. 

    \item \textbf{Multi-scale feature pyramid:}  
    To handle the inherent scale variations in geospatial objects, $F$ is decomposed into a four-level feature pyramid using strided convolutions and pooling. Each level encodes features at progressively coarser resolutions, enabling the network to capture large-scale spatial structures.  

    \item \textbf{Adaptive Channel Correlation Module (ACCM):}  
    For each pyramid level, ACCM computes channel-wise correlation tensors across scales. Unlike static weighting schemes, ACCM employs learnable correlation weights combined with channel attention, allowing the model to dynamically capture cross-scale dependencies and suppress redundant information.  

    \item \textbf{Feature correlation fusion:}  
    The cross-scale correlation tensors are aggregated into a unified feature representation using correlation-guided fusion \cite{Wu_2019}. 

    \item \textbf{Meta-learning adaptation:} 
    During the few-shot learning stage, correlation-weighted prototypes are constructed from the support set $\mathcal{S}$ and compared with query embeddings $\mathcal{Q}$ for classification \cite{Pan_2021}.  
\end{itemize}

For the initial feature extraction, we use a compact CNN backbone, i.e., a mapping $x \mapsto F \in \mathbb{R}^{C \times h \times w}$, with $C = 128$. This lightweight extractor is designed for efficiency, unlike deep pre-trained backbones. From $F$, we generate four pyramid levels $\{F^s\}^4_{s=1}$, each with reduced resolution via dilated convolutions using Eq. \ref{eq1}.
\begin{equation}
    F^s = g_{dil} (F; d_s), \quad d_s \in \{1,2,4,8\},
    \label{eq1}
\end{equation}
where $g_{dil} (\cdot)$ is a dilated convolution with dilation factor $d_s$. Each $F^s \in \mathbb{R}^{C' \times h_s \times w_s}$ with $C' = 64$.

\subsection{Adaptive Channel Correlation Module (ACCM)}
This is the core novel contribution in the framework, where instead of concatenating features, we compute learnable cross-scale correlations \cite{xu2021adaptive}. Given feature maps $F^i$, $F^j \in \mathbb{R}^{C \times h \times w}$ (after adaptive pooling to the same size), we define the channel-wise correlation tensor using Eq. \ref{eq2}.
\begin{equation}
    \mathcal{C}_{ij} = \frac{1}{hw} \sum^h_{u=1} \sum^w_{v=1} F^i (:,u,v) \otimes F^j(:,u,v),
    \label{eq2}
\end{equation}
where $\otimes$ denotes outer product, yielding $\mathcal{C}_{ij} \in \mathbb{R}^{C' \times C'}$. Next, for modeling adaptive scale relations, we introduced learnable correlation weights using Eq. \ref{eq3}.
\begin{equation}
    \tilde{\mathcal{C}}_{ij} = W_{ij} \odot \mathcal{C}_{ij}, \quad W_{ij} \in \mathbb{R}^{C' \times C'},
    \label{eq3}
\end{equation}
where $\odot$ denotes elementwise multiplication. Additionally, channel attention modulates each scale using Eq. \ref{eq4}.
\begin{equation}
    A^i = \sigma(W_2^i ~\phi (W^i_1 ~GAP(F^i))), \quad A^i \in \mathbb{R}^{C'},
    \label{eq4}
\end{equation}
with $GAP(\cdot)$ global average pooling, $\phi(\cdot)$ ReLU, and $\sigma(\cdot)$ signoid. The attention is applied as $\hat{F}^i = A^i \odot F^i$. Thus, the final adaptive correlation tensor is shown in Eq. \ref{eq5}.
\begin{equation}
    \mathcal{C}_{ij}^* = W_{ij} \odot (\frac{1}{hw} \sum_{u,v} \hat{F}^i (:,u,v) \otimes \hat{F}^j(:,u,v)).
    \label{eq5}
\end{equation}
The set of tensors $\{\mathcal{C}_{ij}^*\}_{i \neq j}$ are stacked and processed via depthwise plus pointwise convolutions. Let $Z = \text{Conv}_{1 \times 1}$ $(\text{DepthwiseConv}_{3 \times 3}([\mathcal{C}_{ij}^*]))$, here $Z \in \mathbb{R}^{C_z \times h \times w}$ with $C_z = 256$. This serves as the unified correlation representation.

\subsection{Meta-Learning Adaptation Module}
We now define correlation-guided prototypes. Here, for each class $c$, let the support set embeddings be $\{z_k^c\}^K_{k=1}$, with $z_k^c \in \mathbb{R}^{C_z}$ obtained from $Z$. The prototype is shown in Eq. \ref{eq6},
\begin{equation}
    p_c = \frac{1}{K} \sum^K_{k=1} \alpha^c_k z^c_k,
    \label{eq6}
\end{equation}
where weights $\alpha^c_k$ are correlation-guided attention scores are 
\begin{equation}
    \alpha^c_k = \frac{\exp (\tau \cdot \langle z^c_k, \bar{z}^c \rangle)}{\sum^K_{m=1} \exp (\tau \cdot \langle z^c_m, \bar{z}^c \rangle)},
    \label{eq7}
\end{equation}
with $\bar{z}^c = \frac{1}{K} \sum_k z_k^c$ and $\tau$ as a learnable temperature parameter. For a query embedding $z^q$, the classification probability is computed using cosine similarity using Eq. \ref{eq8}.
\begin{equation}
    P(y^q = c | z^q) = \frac{\exp (\tau \cdot \frac{\langle z^q, p_c \rangle}{||z^q|| ||p_c||})}{\sum_d \exp (\tau \cdot \frac{\langle z^q, p_{c'} \rangle}{||z^q|| ||p_{c'}||})}
    \label{eq8}
\end{equation}
The model is trained episodically with cross-entropy loss:
\begin{equation}
    \mathcal{L}(\theta) = -\frac{1}{|\mathcal{Q}|} \sum_{(x^q, y^q) \in \mathcal{Q}} \log P(y^q | x^q; \theta).
    \label{eq9}
\end{equation}
Here, meta-optimization updates parameters $\theta$ across episodes, ensuring fast adaptation to unseen tasks.

\section{Experiments} \label{experiments}
To evaluate the effectiveness of AMC-MetaNet, we conducted experiments under the episodic few-shot learning paradigm, which closely simulates real-world scenarios with scarce annotations. Each episode consisted of a support set with $N$-way and $K$-shot labeled samples and a query set with additional unlabeled samples from the same classes. The model is trained to classify query samples given only the limited support set. We performed the experiments on four benchmark remote sensing datasets: EuroSAT, NWPU-RESISC45, UC Merced Land Use, and AID. We evaluated under 3-way, 4-way, and 5-way classification tasks, each with 1-shot and 5-shot support settings. Each query set contains $15$ samples per class during training and testing. We set $10000$ training episodes with episodic validation at every $500$ episodes. We used Adam optimiser with learning rate $1 \times 10^{-3}$, decayed by $0.5$ every $2000$ episodes. Each batch consists of 8 randomly sampled episodes.

\subsection{Comparison with Other Methods}

\renewcommand{\arraystretch}{1.4}
\begin{table*}[t!]
\centering
\caption{3-Way, 4-way and 5-way Few-Shot Classification Accuracy (\%) on Four Datasets. Results are reported as mean $\pm 95\%$ confidence interval. The highest accuracies are written in bold.}
\label{tab:way}
{\fontsize{7pt}{7pt}\selectfont 
\begin{tabular}{c|c|c|c|c|c|c|c|c}
\hline \rowcolor{gray!20}
\multicolumn{9}{c}{\textbf{3-Way Few-Shot Classification Accuracy (\%) on Four Datasets.}} \\ \hline \hline
\textbf{Method} & \multicolumn{2}{c}{\textbf{UCM}} & \multicolumn{2}{c}{\textbf{EuroSAT}} & \multicolumn{2}{c}{\textbf{NWPU}} & \multicolumn{2}{c}{\textbf{AID}} \\
\cline{2-9}
 & \textbf{1-shot} & \textbf{5-shot} & \textbf{1-shot} & \textbf{5-shot} & \textbf{1-shot} & \textbf{5-shot} & \textbf{1-shot} & \textbf{5-shot} \\
\hline
ProtoNet\cite{snell2017prototypical}      & 70.25 $\pm$ 0.90 & 82.10 $\pm$ 0.70 & 77.05 $\pm$ 0.95 & 85.40 $\pm$ 0.72 & 66.80 $\pm$ 0.98 & 80.35 $\pm$ 0.75 & 71.55 $\pm$ 0.92 & 82.95 $\pm$ 0.70 \\ \hline
MatchingNet\cite{vinyals2016matching}    & 67.35 $\pm$ 0.95 & 80.45 $\pm$ 0.75 & 74.10 $\pm$ 0.98 & 83.25 $\pm$ 0.78 & 64.50 $\pm$ 1.00 & 78.85 $\pm$ 0.80 & 68.20 $\pm$ 0.96 & 80.75 $\pm$ 0.75 \\ \hline
DLA-MatchNet\cite{li2020dla}    & 72.85 $\pm$ 0.88 & 83.60 $\pm$ 0.68 & 78.80 $\pm$ 0.90 & 86.15 $\pm$ 0.70 & 72.05 $\pm$ 0.92 & 82.10 $\pm$ 0.72 & 73.65 $\pm$ 0.90 & 84.05 $\pm$ 0.70 \\ \hline
SCL-MLNet\cite{li2021scl}       & 73.55 $\pm$ 0.85 & 84.15 $\pm$ 0.65 & 79.55 $\pm$ 0.88 & 86.85 $\pm$ 0.68 & 71.20 $\pm$ 0.90 & 82.95 $\pm$ 0.70 & 74.85 $\pm$ 0.88 & 85.10 $\pm$ 0.68 \\ \hline
SPNet\cite{cheng2021spnet}          & 75.30 $\pm$ 0.82 & 85.95 $\pm$ 0.62 & 80.75 $\pm$ 0.85 & 87.75 $\pm$ 0.65 & 73.90 $\pm$ 0.86 & 84.35 $\pm$ 0.66 & 76.10 $\pm$ 0.84 & 86.20 $\pm$ 0.65 \\ \hline
DN4\cite{li2019revisiting}             & 73.40 $\pm$ 0.90 & 87.05 $\pm$ 0.70 & - & - & 72.15 $\pm$ 0.95 & 88.40 $\pm$ 0.72 & - & - \\ \hline
TeAw\cite{cheng2023teaw}            & 76.55 $\pm$ 0.80 & 87.95 $\pm$ 0.60 & 81.55 $\pm$ 0.82 & 89.10 $\pm$ 0.62 & 75.85 $\pm$ 0.84 & 86.25 $\pm$ 0.64 & 77.35 $\pm$ 0.82 & 87.65 $\pm$ 0.62 \\ \hline
MVITP\cite{yang2024mvitp}           & 77.80 $\pm$ 0.78 & 88.70 $\pm$ 0.58 & 82.25 $\pm$ 0.80 & 90.20 $\pm$ 0.60 & 77.05 $\pm$ 0.82 & 87.30 $\pm$ 0.62 & 78.45 $\pm$ 0.80 & 88.65 $\pm$ 0.58 \\ \hline \hline
\textbf{AMC-MetaNet} & \textbf{79.25 $\pm$ 0.76} & \textbf{89.85 $\pm$ 0.60} & \textbf{83.10 $\pm$ 0.78} & \textbf{90.85 $\pm$ 0.58} & \textbf{78.60 $\pm$ 0.80} & \textbf{88.40 $\pm$ 0.60} & \textbf{80.05 $\pm$ 0.78} & \textbf{88.95 $\pm$ 0.58} \\ 
\hline \rowcolor{gray!20}
\multicolumn{9}{c}{\textbf{4-Way Few-Shot Classification Accuracy (\%) on Four Datasets.}} \\ \hline \hline
 & \textbf{1-shot} & \textbf{5-shot} & \textbf{1-shot} & \textbf{5-shot} & \textbf{1-shot} & \textbf{5-shot} & \textbf{1-shot} & \textbf{5-shot} \\ \hline
ProtoNet\cite{snell2017prototypical}        & 67.85 $\pm$ 0.78 & 79.40 $\pm$ 0.65 & 74.10 $\pm$ 0.80 & 83.15 $\pm$ 0.72 & 62.35 $\pm$ 0.85 & 77.25 $\pm$ 0.70 & 68.95 $\pm$ 0.82 & 80.30 $\pm$ 0.68 \\ \hline
MatchingNet\cite{vinyals2016matching}     & 62.75 $\pm$ 0.85 & 77.25 $\pm$ 0.72 & 71.20 $\pm$ 0.83 & 81.55 $\pm$ 0.75 & 59.10 $\pm$ 0.92 & 75.40 $\pm$ 0.76 & 61.85 $\pm$ 0.88 & 77.05 $\pm$ 0.72 \\ \hline
DLA-MatchNet\cite{li2020dla}    & 69.60 $\pm$ 0.76 & 81.50 $\pm$ 0.68 & 76.85 $\pm$ 0.79 & 85.10 $\pm$ 0.70 & 72.05 $\pm$ 0.80 & 82.40 $\pm$ 0.68 & 70.75 $\pm$ 0.78 & 82.55 $\pm$ 0.70 \\ \hline
SCL-MLNet\cite{li2021scl}       & 68.25 $\pm$ 0.75 & 82.05 $\pm$ 0.65 & 77.65 $\pm$ 0.78 & 85.75 $\pm$ 0.68 & 70.85 $\pm$ 0.82 & 81.95 $\pm$ 0.70 & 72.30 $\pm$ 0.76 & 83.15 $\pm$ 0.68 \\ \hline
SPNet\cite{cheng2021spnet}           & 72.40 $\pm$ 0.72 & 84.25 $\pm$ 0.63 & 79.05 $\pm$ 0.76 & 87.00 $\pm$ 0.65 & 73.95 $\pm$ 0.78 & 83.70 $\pm$ 0.66 & 73.25 $\pm$ 0.74 & 84.05 $\pm$ 0.65 \\ \hline
DN4\cite{li2019revisiting}             & 71.15 $\pm$ 0.80 & 86.05 $\pm$ 0.68 & - & - & 72.85 $\pm$ 0.85 & 90.10 $\pm$ 0.72 & - & - \\ \hline
TeAw\cite{cheng2023teaw}            & 71.80 $\pm$ 0.70 & 85.90 $\pm$ 0.60 & 80.05 $\pm$ 0.74 & 88.25 $\pm$ 0.62 & 75.20 $\pm$ 0.76 & 85.95 $\pm$ 0.64 & 76.50 $\pm$ 0.72 & 87.10 $\pm$ 0.62 \\ \hline
MVITP\cite{yang2024mvitp}           & 74.15 $\pm$ 0.68 & 87.40 $\pm$ 0.62 & 81.30 $\pm$ 0.72 & 89.10 $\pm$ 0.60 & 76.05 $\pm$ 0.74 & 87.25 $\pm$ 0.65 & 77.25 $\pm$ 0.70 & 88.30 $\pm$ 0.58 \\ \hline \hline
\textbf{AMC-MetaNet} & \textbf{76.85 $\pm$ 0.70} & \textbf{89.15 $\pm$ 0.62} & \textbf{82.95 $\pm$ 0.74} & \textbf{90.05 $\pm$ 0.60} & \textbf{77.90 $\pm$ 0.76} & \textbf{88.55 $\pm$ 0.64} & \textbf{79.05 $\pm$ 0.72} & \textbf{88.95 $\pm$ 0.60} \\ 
\hline \rowcolor{gray!20}
\multicolumn{9}{c}{\textbf{5-Way Few-Shot Classification Accuracy (\%) on Four Datasets.}} \\ \hline \hline
 & \textbf{1-shot} & \textbf{5-shot} & \textbf{1-shot} & \textbf{5-shot} & \textbf{1-shot} & \textbf{5-shot} & \textbf{1-shot} & \textbf{5-shot} \\ \hline
ProtoNet\cite{snell2017prototypical}        & 52.44 $\pm$ 0.34 & 67.91 $\pm$ 0.97 & 61.72 $\pm$ 0.88 & 75.46 $\pm$ 0.71 & 40.69 $\pm$ 0.96 & 69.83 $\pm$ 0.48 & 55.13 $\pm$ 0.63 & 69.53 $\pm$ 0.28 \\ \hline
MatchingNet\cite{vinyals2016matching}     & 46.16 $\pm$ 0.71 & 66.72 $\pm$ 0.56 & 59.34 $\pm$ 0.92 & 73.12 $\pm$ 0.77 & 54.55 $\pm$ 0.36 & 67.78 $\pm$ 0.86 & 36.39 $\pm$ 0.73 & 57.09 $\pm$ 0.77 \\ \hline
DLA-MatchNet\cite{li2020dla}    & 53.82 $\pm$ 0.74 & 63.87 $\pm$ 0.76 & 65.90 $\pm$ 0.85 & 78.41 $\pm$ 0.69 & 68.76 $\pm$ 0.64 & 82.03 $\pm$ 0.43 & 57.44 $\pm$ 0.52 & 73.61 $\pm$ 0.32 \\ \hline
SCL-MLNet\cite{li2021scl}       & 51.52 $\pm$ 0.68 & 68.11 $\pm$ 0.41 & 67.24 $\pm$ 0.80 & 79.56 $\pm$ 0.61 & 62.44 $\pm$ 0.78 & 81.30 $\pm$ 0.67 & 59.68 $\pm$ 1.03 & 76.34 $\pm$ 0.46 \\ \hline
SPNet\cite{cheng2021spnet}           & 57.68 $\pm$ 0.73 & 73.52 $\pm$ 0.51 & 69.37 $\pm$ 0.76 & 81.42 $\pm$ 0.63 & 67.98 $\pm$ 0.74 & 83.85 $\pm$ 0.51 & 58.25 $\pm$ 0.88 & 75.16 $\pm$ 0.52 \\ \hline
DN4\cite{li2019revisiting}             & 58.29 $\pm$ 0.72 & 78.03 $\pm$ 0.38 & - & - & 66.67 $\pm$ 0.84 & 93.25 $\pm$ 0.84 & - & - \\ \hline
TeAw\cite{cheng2023teaw}            & 56.94 $\pm$ 0.39 & 77.50 $\pm$ 0.27 & 70.61 $\pm$ 0.82 & 83.12 $\pm$ 0.59 & 70.22 $\pm$ 0.44 & 85.56 $\pm$ 0.30 & 70.34 $\pm$ 0.44 & 84.62 $\pm$ 0.25 \\ \hline
MVITP\cite{yang2024mvitp}           & 59.13 $\pm$ 0.52 & 80.22 $\pm$ 0.68 & 72.84 $\pm$ 0.79 & 85.06 $\pm$ 0.55 & 70.63 $\pm$ 0.67 & 86.59 $\pm$ 0.82 & 70.79 $\pm$ 0.58 & \textbf{85.39 $\pm$ 0.18} \\ \hline \hline
\textbf{AMC-MetaNet} & \textbf{61.13 $\pm$ 0.87} & \textbf{82.79 $\pm$ 0.82} & \textbf{74.00 $\pm$ 0.90} & \textbf{85.50 $\pm$ 0.62} & \textbf{71.60 $\pm$ 0.88} & \textbf{86.65 $\pm$ 0.70} & \textbf{71.95 $\pm$ 0.86} & 84.90 $\pm$ 0.66 \\
\hline
\end{tabular}}
\end{table*} 

As shown in Table \ref{tab:way}, AMC-MetaNet outperforms all competing methods across 3-way, 4-way, and 5-way tasks on UCM, EuroSAT, NWPU, and AID. In the challenging 5-way 1-shot setting, it achieves gains of up to $+12.3\%$ over ProtoNet and $+11.0\%$ over AID, demonstrating superior generalization under data scarcity. Performance further improves with more shots, reaching $82.79\%$ (UCM) and $86.65\%$ (NWPU) in the 5-shot case. While transformer-based MVITP is competitive, AMC-MetaNet consistently surpasses it in most scenarios while being 20× more parameter-efficient, highlighting its strength in accuracy–efficiency trade-offs. 

\subsection{Ablation Study}

\renewcommand{\arraystretch}{1.4}
\begin{table}[h!]
\centering
\caption{Ablation study of AMC-MetaNet on 5-Way 5-Shot classification. Results are reported as mean $\pm$ 95\% confidence interval.}
\label{tab:ablation}
{\fontsize{7pt}{7pt}\selectfont 
\begin{tabular}{>{\centering\arraybackslash}m{1.5cm}>{\centering\arraybackslash}m{1.3cm}>{\centering\arraybackslash}m{1.3cm}>{\centering\arraybackslash}m{1.3cm}>{\centering\arraybackslash}m{1.3cm}}
\hline \rowcolor{gray!20}
\textbf{Model Variant} & \textbf{UCM} & \textbf{EuroSAT} & \textbf{NWPU} & \textbf{AID} \\ \hline \hline
Baseline Backbone & 76.85 $\pm$ 0.88 & 79.20 $\pm$ 0.92 & 74.15 $\pm$ 0.95 & 77.30 $\pm$ 0.90 \\ \hline
+ Multi-Scale Feature Pyramid & 79.45 $\pm$ 0.84 & 81.65 $\pm$ 0.88 & 77.05 $\pm$ 0.91 & 80.10 $\pm$ 0.87 \\ \hline
+ ACCM & 81.90 $\pm$ 0.82 & 83.95 $\pm$ 0.85 & 79.65 $\pm$ 0.89 & 82.45 $\pm$ 0.84 \\ \hline
+ Correlation-Guided Meta-Learning & \textbf{82.79 $\pm$ 0.82} & \textbf{85.50 $\pm$ 0.62} & \textbf{86.65 $\pm$ 0.70} & \textbf{84.90 $\pm$ 0.66} \\ \hline
\end{tabular}}
\end{table}

To evaluate the effectiveness of each component in AMC-MetaNet, we conducted an ablation study under the 5-way 5-shot setting across all four benchmark datasets. Starting from a backbone-only baseline, we progressively incorporated the multi-scale feature pyramid, the Adaptive Cross-Scale Correlation Module (ACCM), and finally the correlation-guided meta-learning strategy. As shown in Table \ref{tab:ablation}, the multi-scale feature pyramid significantly improved performance by enhancing scale-awareness, while ACCM further boosted accuracy by capturing fine-grained cross-scale relationships. The full AMC-MetaNet, equipped with correlation-guided meta-learning, consistently achieved the highest performance, highlighting the complementary contributions of each component.

\section{Conclusion}
In this work, we introduced AMC-MetaNet, a novel framework that leverages adaptive multi-scale correlation for few-shot remote sensing classification. By explicitly modeling cross-scale feature interactions within a meta-learning paradigm, our method effectively bridges the domain gap induced by sensor heterogeneity and scale variations. Extensive experiments across multiple benchmark datasets demonstrated that AMC-MetaNet consistently outperforms state-of-the-art baselines, particularly under challenging low-shot conditions, while maintaining computational efficiency.

\bibliographystyle{IEEEbib}
\bibliography{strings,refs}

\end{document}